\documentclass[10pt,twocolumn,letterpaper]{article}
\pdfoutput=1
\usepackage{authblk}

\usepackage[pagenumbers]{cvpr}
\usepackage{multirow}
\usepackage{array}
\usepackage{bbm}
\usepackage{graphicx}
\usepackage{amsmath}
\usepackage{amssymb}
\usepackage{booktabs}
\usepackage{bbding}
\usepackage{color}
\usepackage{appendix}
\usepackage{fancyhdr}
\usepackage{CJKutf8}
\usepackage[dvipsnames]{xcolor}

\definecolor{cvprblue}{rgb}{0.21,0.49,0.74}
\usepackage[pagebackref,breaklinks,colorlinks,citecolor=cvprblue]{hyperref}
\usepackage{xcolor}
\definecolor{DarkGreen}{rgb}{0.43, 0.68, 0.28}

\newcommand\keywords[1]{\textbf{Keywords}: #1}

\title{Gaussian Control with Hierarchical Semantic Graphs in 3D Human Recovery}

\author[1,2]{Hongsheng Wang}
\author[2]{Weiyue Zhang}
\author[2]{Sihao Liu}
\author[2]{Xinrui Zhou}
\author[2]{Jing Li}
\author[2]{Zhanyun Tang}
\author[1]{\\ Shengyu Zhang\textsuperscript{$\dag$}} 
\author[1]{Fei Wu}
\author[2]{Feng Lin}

\affil[1]{Zhejiang University, China}
\affil[2]{Zhejiang Lab, China}

\makeatletter
\renewcommand\AB@affilsepx{, \protect\Affilfont}
\makeatother

\begin{document}

\maketitle

\begin{CJK}{UTF8}{gbsn}
\renewcommand{\thefootnote}{\fnsymbol{footnote}}
\footnotetext[2]{Corresponding Author.}
\renewcommand\thefootnote{}
\footnote{This work has been submitted to the IEEE for possible publication. Copyright may be transferred without notice, after which this version may no longer be accessible.}
\begin{abstract}

Although 3D Gaussian Splatting (3DGS) has recently made progress in 3D human reconstruction, it overlooked blurred details in the reconstruction of the human body, particularly at the junctions and surface features. To address this gap, we introduce the \textbf{H}ierarchical Graph H\textbf{u}man \textbf{G}au\textbf{s}sian Control (HUGS) framework for achieving high-fidelity 3D human reconstruction. Our method enhances the learning of relationships in motion  at junctions of the human body by innovatively establishing semantic-level constraints. Additionally, we improve the learning of internal appearance connections within different parts of the human body by extracting high-frequency information between Gaussian points with identical semantic labels, thereby reconstructing high-frequency details within each part. Extensive experiments demonstrate that our method excels in human body reconstruction, especially at junctions and surface features.Codes are available at \href{https://wanghongsheng01.github.io/HUGS/}{https://wanghongsheng01.github.io/HUGS/}.

\end{abstract}


\section{Introduction}

\begin{figure*}[htbp]
    \centering
    \includegraphics[width=0.8\linewidth]{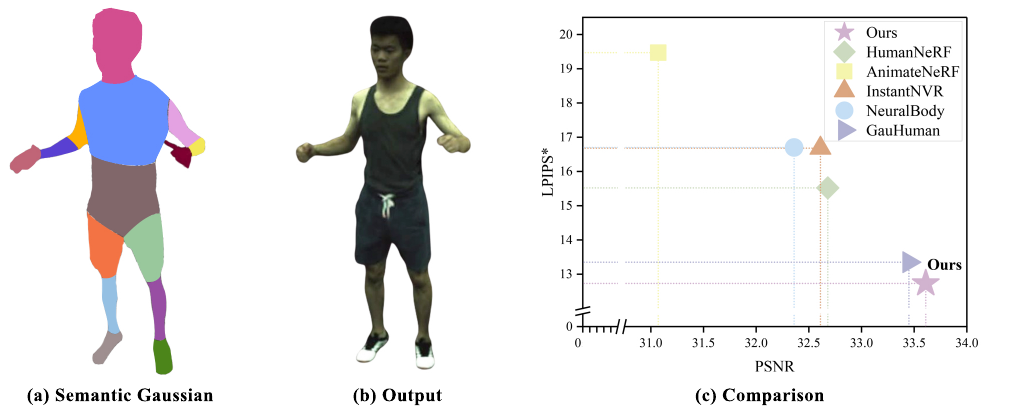}
    \caption{Human Gaussian Control with Hierarchical Semantic Graphs. (a) is a human Gaussian point cloud with semantic labels. (b) is the rendering output of (a). (c) is the result of our method compared with other methods on the Monocap dataset. LPIPS* = LPIPS $\times$ 1000.}
    \label{figure:over_view}
\end{figure*}
With the development of technologies such as computer graphics, virtual reality, and artificial intelligence, the generation of articulate 3D digital humans has become a popular research field. It holds immense potential value in industries like holographic transmission, and the metaverse. Traditional 3D representation methods~\cite{mesh,overview,point,restruct1,restruct2,restruct3,restruct4} like meshes and point clouds require dense views for human reconstruction~\cite{lsh1,lsh2,lsh3,lsh4,lsh5,lsh6,lsh7,lsh8}, limiting their applications in sparse views for human reconstruction. 

Current research incorporates the SMPL model into the 3DGS process~\cite{3dgs,zhang2020learning,jiang2024uv,li2024gaussianbody}. However, these methods still suffer from blurred details in the reconstruction of the human body, particularly at the junctions and surface features. These issues arise because existing approaches only inherit positional information from the SMPL model while neglecting the inherent connections between different body parts. Additionally, in monocular dynamic scenarios, the existing 2D pixel-level supervision often smooths out 3D discrepancies during complex motion deformations, failing to deeply capture the intricate relationships between local Gaussian points within a region. This leads to the blurring of important details such as clothing wrinkles and muscle textures.

To bridge the gap, we propose \textbf{Hu}man \textbf{G}aussian Control with Hierarchical \textbf{S}emantic Graphs (\textbf{HUGS}), to deeply learn the relationships between Gaussian points both at the connections of different body parts and within body parts. To address the blurring issues at body part junctions in 3D reconstruction, we introduce the Inter-Semantic Kinematic Topology module, which improves the learning of relationships in motion at these junctions. In this module, we innovatively inherit semantic information from SMPL into Gaussian points to establish semantic-level constraints. Additionally, we construct a 3D Gaussian point graph structure and employ a random walk method to provide positional embedding for Gaussian points, thereby creating contrastive learning samples based on the prior human topological Hierarchical structure. 
Meanwhile, we propose the Intro-Surface Disentanglement module to handle the blurriness in every internal body part. By extracting high-frequency information between Gaussian points with the same semantic label based on a graph structure, we subsequently increase the density of Gaussian points at these high-frequency locations.

Our main contributions can be summarized as follows:
\begin{enumerate}
    \item We propose an Inter-Semantic Kinematic Topology method based on a topological graph of Gaussian point semantic relationships. This method addresses the blurring issues at the connections of different body parts in 3D reconstruction by learning the relationships in motion at the junctions.
    \item We introduce an Intro-Surface Disentanglement method based on high-frequency extraction between Gaussian points with the same semantics, structured as a graph. This approach handles the internal blurriness within body parts by learning the relationships in appearance.
    \item To validate the effectiveness of our method, we conduct extensive experiments and comparisons. Compared to existing methods, the results demonstrate significant improvements in rendering quality and body structure, especially at the junctions and surface features.
\end{enumerate}

\section{Related Work}
\subsection{Statistical Parametric Human Body Model}
The Statistical Parametric Human Body Model (SMPL) is a parametric model capable of accurately representing human body shape and pose\cite{loper2023smpl}. One approach utilized SMPL for body pose and shape recovery, leveraging the optimization results generated by SMPLify as supervision signals to train an end-to-end network, thereby enhancing the realism of the reconstruction results\cite{bogo2016keep}. Building on this, another approach trained 91 keypoint detectors. Subsequently, they optimized the SMPL model parameters to fit these keypoints and further proposed a random forest regression method to directly regress these parameters, albeit at the expense of accuracy\cite{lassner2017unite}. Some researchers employed synthesized data rendered to train a fully convolutional model for depth and body part segmentation\cite{varol2017learning}. In another work, 2D and 3D joint positions were estimated by fitting a bound skeleton model. While 3D rotations could be recovered post-optimization, this approach directly outputs rotations and surface vertices from images\cite{mehta2017vnect}. Another method regressed SMPL pose, shape, and camera parameters, using synthetic data for supervised training and unsupervised fine-tuning during inference, with 2D joints, contours, and motion losses.\cite{tung2017self}.One study regressed SMPL and camera parameters and applied adversarial training, where the predicted results were input to a discriminator network for real-vs-fake classification\cite{kanazawa2018end}. A multi-task approach was proposed to estimate 2D/3D pose, pixel segmentation, and volumetric shape, utilizing real volumetric data generated by SMPL, but without embedding the SMPL function within the network.\cite{varol2018bodynet}. A deep learning method named SMPLR was subsequently introduced. It accurately recovers 3D human pose and shape from RGB images by estimating 3D joints and sparse landmarks, followed by the parameter regression\cite{madadi2020smplr}.Previous studies utilizing SMPL for human body reconstruction overlooked the relationships in motion at the junctions, leading to blurring issues at these connection points. To address this, we propose transferring the semantic information from SMPL to Gaussian points. This approach will establish semantic-level constraints, effectively resolving the blurring issues at the junctions during reconstruction.
\subsection{3D Gaussian Human}
3D Gaussian Splatting\cite{3dgs} is an explicit scene representation that technique utilizing flexible Gaussian ellipsoids, enabling more versatile 3D scene reconstruction and faster rendering compared to NeRF. Human Gaussian Splatting\cite{HUGSRT} has showcased 3D Gaussian Splatting as an efficient alternative to NeRF. This study focuses on the application of 3D Gaussian Splatting in human body reconstruction\cite{gauhuman,humangaussian,human101, animaGaussian, HUGS, HUGSRT}. While extracting the human body from multi-view videos using 3D Gaussian Splatting is feasible\cite{HUGSRT,animaGaussian}, challenges persist in reconstructing high-frequency regions within various parts of the human body with high quality\cite{HUGSRT}. The randomness in Gaussian generation during reconstruction can hinder the extraction of 3D human body mesh information\cite{gauhuman} and lead to an excessive reliance on precise pose inputs\cite{animaGaussian}. Insufficient human body geometry information may result in parts not properly following human movements (e.g., cloth deformation with poses)\cite{HUGSRT}. We propose that these blurry details within parts of the human body can be mitigated through supervised Gaussian generation that incorporates human semantics. By extracting high-frequency information between Gaussian points with the same semantic label, we can learn the internal relationships in appearance within various parts of the human body\cite{gauhuman}.

\section{Methods}
We propose \textbf{Hu}man \textbf{G}aussian Control with Hierarchical \textbf{S}emantic Graphs (\textbf{HUGS}) to generate Gaussian humans, ensuring both junctions of human body and realistic human appearance. To learn the semantic and motion association of body parts, we introduce Inter-Semantic Kinematic Topology in Section \ref{sec:3.1}, aligning human semantic and kinematic topology with 3D humans to capture complex geometric features of body parts and kinematic correlations. To learn relationships in appearance within body parts, we introduce Intro-Surface Disentanglement in Section \ref{sec:3.2}, which disentangles high-frequency features from human body features in each body part, refining the local structures of significant discrepancies on the human surface.
\begin{figure*}
    \centering
    \includegraphics[width=0.8\textwidth]{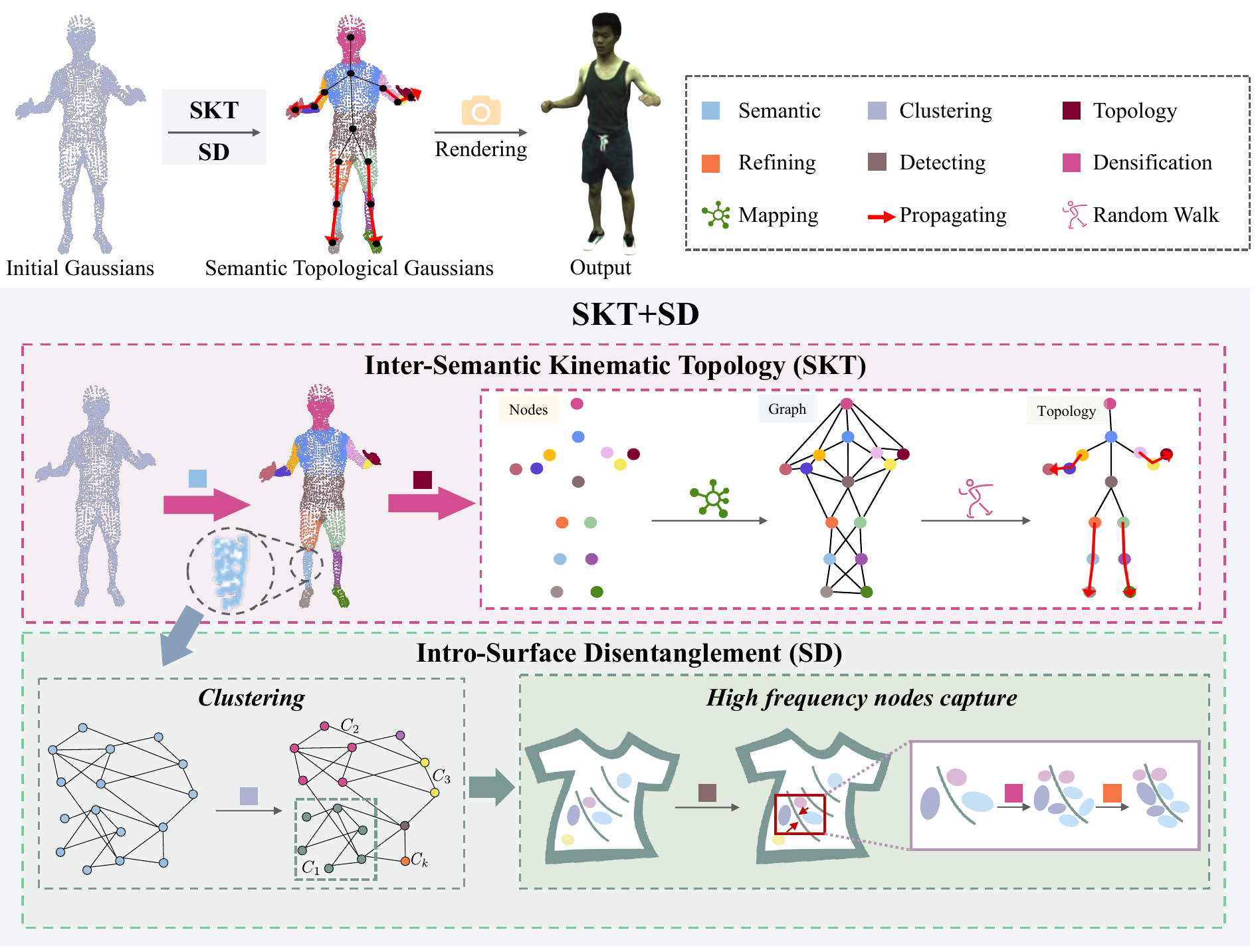}
    \caption{\textbf{HUGS framework.} We introduce \textbf{Hu}man \textbf{G}aussian Control with Hierarchical \textbf{S}emantic Graphs (\textbf{HUGS}) as a method for generating Gaussian humans, ensuring both realistic human appearance and anatomical structure. The input initialized point cloud is mapped using the \textbf{SKT} module to establish a graph structure that encodes the semantic topological relationships between different body parts. High-frequency regions are identified by \textbf{SD} for each body part, and the density of Gaussian points is increased in these areas. Finally, the adjusted Gaussian points are rendered to produce the image output.}
    \label{Figure:enter-label}
\end{figure*}
\subsection{Preliminary}
\label{sec:3.1}
\textbf{SMPL~\cite{smpl}}: 
The SMPL model is a pre-trained parametric human model representing body shape and pose. In SMPL, body shape and pose are controlled by two main parameters: $\beta$ and $\theta$, where $\beta$ represents 10 parameters for proportions such as height, weight, and head-to-body ratio, and $\theta$ represents 75 parameters for overall body motion posture and relative angles of 24 joints. These parameters enable the SMPL model to generate various human body shapes and poses.

\noindent\textbf{3D Gaussian Splatting~\cite{3dgs}}: 3D Gaussian Splatting is a technique used in computer graphics and visualization to represent and render 3D data. It is a method of approximating a continuous volume distribution by discretizing it into Gaussian points and assigning each pixel a value based on the contribution of nearby Gaussian points. Each Gaussian point is defined by its position center $p$ and a full 3D covariance matrix $\Sigma$. The formula is: 
\begin{equation}
  G(\boldsymbol{x})=\frac{1}{(2 \pi)^{\frac{3}{2}}|\boldsymbol{\Sigma}|^{\frac{1}{2}}} e^{-\frac{1}{2}(\boldsymbol{x}-\boldsymbol{p})^{T} \boldsymbol{\Sigma}^{-1}(\boldsymbol{x}-\boldsymbol{p})}.   
\end{equation}

To ensure the positive semi-definiteness of $\Sigma$, it is decomposed $\Sigma$ into two learnable parts: a quaternion representing rotation $r \in R^4$ and a vector representing scaling $s \in R^3$. By transforming these parts into the corresponding rotation matrix $R$ and scaling matrix $S$, we can obtain the covariance matrix $\Sigma = RSS^T R^T$.
We calculate the Gaussian distributions overlapping with each pixel and compute their color $c_i$ and density contribution $\alpha_i$. The color of each pixel can be obtained by blending N-ordered Gaussians, according to the formula:
\begin{equation}
    \hat{C}=\sum_{i \in N} c_{i} \alpha_{i} \prod_{j=1}^{i-1}(1-\alpha_{i}).
    \label{eq:2}
\end{equation}

\subsection{Inter-Semantic Kinematic Topology}
\label{sec:3.2}
In this section, we introduce our proposed Inter-Semantic Kinematic Topology Aware method, which is based on a topological graph of Gaussian point semantic relationships. This method comprises two submodules: Semantic Alignment Gaussian and Inter-Joint Graph. By learning the motion relationships at the connections of different body parts, it addresses the blurring issues at these junctions in 3D reconstructions.

Applying 3D Gaussian Splatting directly to human reconstruction presents several challenges. Relying on 2D pixel-level supervision to refine the initial coarse Gaussian points distribution, this approach lacks explicit learning of Gaussian points based on different body parts of the human anatomy and overlooks the topological relationships between these parts. The human complex motion may bring about local structures being occluded, generating semantic confusion in Gaussian points at the junction of body parts. Gaussian points fail to correctly match their corresponding spatial positions in the human body, resulting in geometric distortion in the occluded parts. Therefore, considering the intrinsic geometric structures of different body parts and their topological relationships is crucial in addressing the reconstruction problem of local geometric distortion in the occluded parts. 

We explore following the human kinematic tree to guide Gaussian points optimization, thus better understanding the intrinsic geometric structures of different body parts and their topological relationships. The core idea of this approach is to model the human as a topological structure containing semantic information, where each body part is endowed with specific kinematic properties and geometric features. 

Traditional 3D Gaussian parameters include the 3D center position (mean), 3D size (anisotropic covariance), opacity, and color (spherical harmonics coefficients). These parameters are differentiable and can be easily rendered onto a 2D plane. To inject semantic information of the body parts into the 3D Gaussian optimization process, we introduce a new parameter for each Gaussian point, namely the semantic attribute. We initialize the Gaussian points of the human using the SMPL model~\cite{smpl} and then initialize the body semantic attribute for each Gaussian point using the semantic labels of the SMPL. During training, we dynamically track the specific body parts associated with each Gaussian point. During densification, new Gaussian points inherit the body semantic attribute from their parent nodes. By explicitly indicating specific body part semantics in 3D space, human body semantic labels address the issue of semantic confusion caused by occlusion at the junction of body parts. 

\textbf{Semantic Alignment Gaussian}: The semantic attributes of Gaussian points explicitly inject semantics into the 3D Gaussian optimization process. Utilizing these semantic labels, we establish connections between 2D and 3D spaces. In the 3D space, we expect points that are close in distance to have similar semantic attributes. When rendered onto a 2D plane, these semantic attributes are also projected. We use a pre-trained CLIP model to establish correspondences between semantically segmented 2D supervisory images and the rendered semantic projection masks. In this way, we can better associate Gaussian points belonging to the same part of the body, enrich the 2D supervisory information, and strengthen the connection between the 3D space and 2D supervision. The loss function formula corresponding to semantic consistency constraint is shown as follows:
\begin{equation}
    \mathcal{L}_{\text{semantic}} =\frac
{1}{N}\sum_{i=1}^{N}(D(\mathcal{X}_i,y_i)+\frac{1}{k}\sum_{j \in Q_i}D(\mathcal{X}_j,y_i)),
\label{eq:3}
\end{equation}
where $N$ represents the total number of Gaussian points, $y_i$ represents the semantic attribute of Gaussian point $i$, $Q_i$ contains the $k$ nearest neighbors of Gaussian point $i$ in 3D space. $\mathcal{X}_i$ represents the semantic information of the pixel rendered by Gaussian point $i$, 3DGS employs point-based rendering techniques ($\alpha$-blending) to render the color attributes of 3D Gaussian points onto the 2D-pixel plane, function $D$($\cdot$) is used to determine whether the Gaussian point semantic properties are consistent with the true semantics corresponding to the pixels to which they are mapped. Similarly, we blend the semantic attributes of the human body from 3D space to a 2D-pixel plane by mixing them in the order of the influence of on pixels.The rendering formula is as follows:
\begin{equation}
\mathcal{X}_i = \sum_{p \in \mathcal{N}} y_{p} \alpha_{p} \prod_{j=1}^{p-1}\left(1-\alpha_{j}\right).
\end{equation}

\noindent where $\mathcal{X}_i$ represents the semantic labels of pixel $i$, which are derived from Gaussian point semantic attribute $\alpha$-blending(Equation~\ref{eq:2}), $\mathcal{N}$ represents the set of all Gaussian points that overlap with the pixels rendered by Gaussian point $i$, $y_p$ represents the semantic attribute corresponding to the 3D Gaussian point $p$, and $\alpha_p$ represents the influence factor of the Gaussian point $p$ for rendering pixels.

\textbf{Inter-Joint Graph}
Although the semantic consistency constraint can independently optimize the geometric structure of each body part, the different parts of the human body are not completely independent. There are mutual influences and collaborative effects between various body parts. For example, the position and posture of the arms are influenced by the positions and postures of the palms and shoulders. However, due to the complexity of human motion, the linkage effects between body parts cannot be explicitly expressed. Therefore, we explore using the explicit topological structure of the human body to learn the implicit linkage effects between Gaussian points of different body parts. To achieve this goal, we introduce topological coherence constraints to learn the correlations between different body parts. To incorporate topological information into the human reconstruction process, we use the spatial position information of Gaussian points to create a topological graph for the human body, where each Gaussian point corresponds to a node in the graph. For each node, we calculate the $k$ (same in Equation~\ref{eq:3}) nearest neighbors based on spatial distance and add edges, with the edge weights represented by the distances. In the topological graph, we use random walk algorithms~\cite{node2vec, deepwalk, struc2vec, random} to propagate information and sample, generating positional embedding vector for each node to capture the topological associations between different Gaussian points. Additionally, we establish a standard prior topological relationship graph for 15 key regions of the human body based on segmentation and joint associations as supervisory information, where each key part is associated with the two adjacent parts in the human body topology diagram. We then sample from all point clouds to construct contrastive learning samples. If the semantic labels of the two sampled points match the prior associations, they are considered positive samples; otherwise, they are negative samples. The loss function corresponding to the topological coherence constraint is as follows:

\begin{equation}
    \mathcal{L}_{\text {topology }}= \frac{1}{MN} \sum_{i=1}^{N}\sum_{t_j}^{M} \theta(t_i, t_j^-) - \theta(t_i, t_j^+).
\end{equation}

\noindent where $N$ represents the total number of Gaussian points, $M$ represents the number of positive and negative samples used for contrast learning, $t_i$ represents the embedding vector for node $i$, which contains topological information obtained by the random walk algorithm,
$t_j^+$ represents the positive samples from body parts that have a linkage effect with the body part corresponding to node $i$, such as Arm and Hand,  $t_j^-$ is the negative sample from body parts that do not have a linkage effect with the body part corresponding to node $i$.

\begin{table*}[ht]
\centering
\caption{Quantitative comparison of our method and other baseline methods on the ZJU-MoCap and MonoCap datasets. We use bold font to highlight the best result and underline the second-best result of each metric. Our method achieves the best PSNR and LPIPS on both datasets. LPIPS* = 1000 × LPIPS.}
\tabcolsep=10pt
\resizebox{0.8\linewidth}{!}{
\begin{tabular}{l|c c c |c c c}
\toprule
\multicolumn{1}{c|}{\multirow{2}{*}{Method}} & \multicolumn{3}{c|}{ZJU-Mocap} & \multicolumn{3}{c}{MonoCap}

       \\
              & PSNR↑     & SSIM↑   & LPIPS*↓ & PSNR↑  & SSIM↑ & LPIPS*↓ \\
\midrule
NeuralBody    & 29.03     & 0.964    & 42.47   & 32.36  & 0.986  & 16.7  \\
HumanNeRF     & 30.66     & 0.969    & 33.38   & 32.68  & 0.987  & 15.52 \\
AnimateNeRF   & 29.77     & 0.965    & 46.89   & 31.07  & 0.985  & 19.47 \\
InstantNVR    & 31.01     & \textbf{0.971} & 38.45  & 32.61  &\textbf{0.988} &  16.68 \\
InstantAvatar & 29.73     & 0.938    & 68.41   & 30.79  & 0.964  & 39.75 \\
GauHuman      & \underline{31.34}     & 0.965    & \underline{30.51}   & \underline{33.45}  & 0.985  & \underline{13.35} \\
Ours          & \textbf{31.35}  & 0.964	& \textbf{28.93}	& \textbf{33.61}	& 0.984	& \textbf{12.73} \\
\bottomrule
\end{tabular}}
\label{tab:comparison}
\end{table*}

Based on Semantic Kinematic Topology, we initially express both the geometric structure of the body parts and the topological correlations between them. However, we observed fuzzy geometric shapes in the local structures with significant discrepancies on the human surface, such as clothing wrinkles and muscle textures. In the next subsection, we will introduce the Intro-Surface Disentanglement module designed to address this issue.

\subsection{Intro-Surface Disentanglement}

In this section, we present our proposed Intro-Surface Disentanglement method, which is based on the high-frequency extraction between Gaussian points with the same semantics, structured as a graph. This method aims to address the internal blurriness within body parts in 3D reconstructions by learning the appearance relationships within these parts.

Based on Semantic Kinematic Topology, we initially express both the geometric structure of the body parts and the topological correlations between them. However, we observe fuzzy geometric shapes in the local structures with significant discrepancies on the human surface, such as clothing wrinkles and muscle textures. These discrepancies are primarily found in high-frequency regions of the human surface. Therefore, capturing high-frequency features becomes essential to guide the optimization of local structures with significant differences. As a result, we explore disentangling high-frequency features from the global human features and utilize them to refine Gaussian point density and distribution corresponding to the local structures with significant discrepancies. 

To address this challenge, we introduce a surface decoupling module. High-frequency signals manifest as anomalous nodes in the human topological graph, which are significantly different from adjacent nodes. First, We cluster all the Gaussian points based on their semantic label vectors. In each iteration, we identify specific nodes as candidate high-frequency nodes by evaluating the average magnitude of structural differences between the selected node and all nodes in the same cluster. Ultimately, we select the nodes with the highest average magnitude of structural differences within each cluster as high-frequency nodes. The algorithm formula is as follows:

\begin{equation}
     H_m = \mathop{\mathrm{argmax}}\limits_{i \in C_m} \frac{1}{C_m - 1} \sum_{j \in C_m, j \neq i} \text{similar} \left(A_i,A_j\right).
\end{equation}

\noindent where $A_i$ represents the basic attribute of Gaussian points (color, opacity, and radii), $C_m$ represents the set of all points with label m. $\text{similar}(\cdot)$ is used to evaluate the similarity between two nodes.

The positions of Gaussian points represented by high-frequency nodes are considered as the cluster centroids of local structures with significant discrepancies. To better capture and express these local structures of significant discrepancies, we perform densification operations on these Gaussian points, enhancing the local rendering granularity to focus on guiding the density and attribute optimization of Gaussian points in these areas. This approach aims to better represent the local structures of significant discrepancies on the human surface.

\subsection{Semantic-Awareness Optimization}
Different from random or Structure-from-Motion (SfM) initialization methods for Gaussian point clouds, we directly sample 6890 point clouds from the SMPL model as initialization, reducing the cost of network training. During the inference stage, we use the optimized LBS weights and pose parameters saved during the training stage, which not only maintains high rendering quality but also improves modeling speed.
The final loss function combines image generation quality, semantic consistency constraint, and topological coherence constraint. The loss function is as follows:
\begin{equation}
    \mathcal{L} = \lambda_1 \mathcal{L}_{image} +\lambda_2\mathcal{L}_{semantic}+ \lambda_3\mathcal{L}_{topology}.
\end{equation}

\section{Experiments}

\subsection{Datasets and Metrics}
\noindent\textbf{ZJU-MoCap Dataset.} The ZJU-Mocap dataset~\cite{NeuralBody_ZJU-Mocap} is a prominent benchmark in human modeling from videos. Similar to previous work~\cite{InstantNVR,HumanNeRF}, we select 6 human subjects (377, 386, 387, 392, 393, 394) from the dataset to conduct experiments. 
Following~\cite{InstantNVR}, we use one camera for training and the remaining cameras for evaluation. For each subject, we sample 1 frame every 5 frames and collect 100 frames for training.

\noindent\textbf{MonoCap Dataset.} The MonoCap Dataset includes four multi-view videos from DeepCap~\cite{DeepCap} and DynaCap~\cite{DynaCap} datasets, collected by~\cite{Monocap_AS}. This dataset provides essential details like camera parameters, SMPL parameters, and human masks. As in~\cite{Monocap_AS}, we use one camera view for training and selected ten uniformly distributed cameras for testing. For each subject, we sample 1 frame every 5 frames and collect 100 frames for training.

\noindent\textbf{Evaluation Metrics.} We use PSNR, SSIM, and LPIPS as quality evaluation metrics. The average results of the 6 sub-datasets of the ZJU-MoCap dataset are used as the experimental results for each method on this dataset. Similarly, the average results of the 4 sub-datasets of the MonoCap dataset are used as the experimental results for each method on this dataset.

\subsection{Quantitative Results}

To verify the effectiveness of our method in solving the geometric distortion problem in reconstructing the human body, we compare our method with NeuralBody~\cite{NeuralBody_ZJU-Mocap}, HumanNeRF~\cite{HumanNeRF} AnimateNeRF~\cite{AnimaNerf}, InstantNVR~\cite{InstantNVR} InstantAvatar~\cite{InstantAvatar} GauHuman~\cite{gauhuman} on the ZJU dataset and the Monocap dataset, as shown in Table~\ref{tab:comparison}. \textbf{Compared with previous methods, our model achieve the best results on both PSNR and LPIPS on both datasets, outperforming previous methods.}

As implicit representations, HumanNeRF~\cite{HumanNeRF} and InstantNVR~\cite{InstantNVR} ignore the geometric structure of human body parts, which can easily confuse the information of body parts and lead to structural distortion. \textbf{Our method independently optimizes the specific geometric structure of each body part, using semantic information to constrain the supervision of Gaussian point clouds.} The 3DGS method, Gauhuman~\cite{gauhuman}, introduces the SMPL model as a prior. However, it only uses static human information during the initialization phase, still challenging in handling the deformation caused by complex human motion. Our model captures implicit association between different body parts, thereby constraining the topological structure of Gaussian points with the kinematic correlation to model the motion architecture. The improvement of our model's LPIPS metric on the ZJU and Monocap datasets proves the effectiveness of our method.

In terms of local details of the human body, both InstantNVR~\cite{InstantNVR} and Gahuman~\cite{gauhuman} do not pay special attention to the high-frequency features of the human body surface, which has a huge impact on the quality of the reconstructed human body surface. The Surface Distinction module in our model extracts Gaussian points that express high-frequency features from the topology graph and performs densification operations to fit the region of the body surface with high-frequency characteristics. \textbf{The improvement of PSNR proves that our model is effective in restoring high-frequency information.}

\subsection{Qualitative Results}

\begin{figure*}[t]
    \centering
    \includegraphics[width=0.65\linewidth]{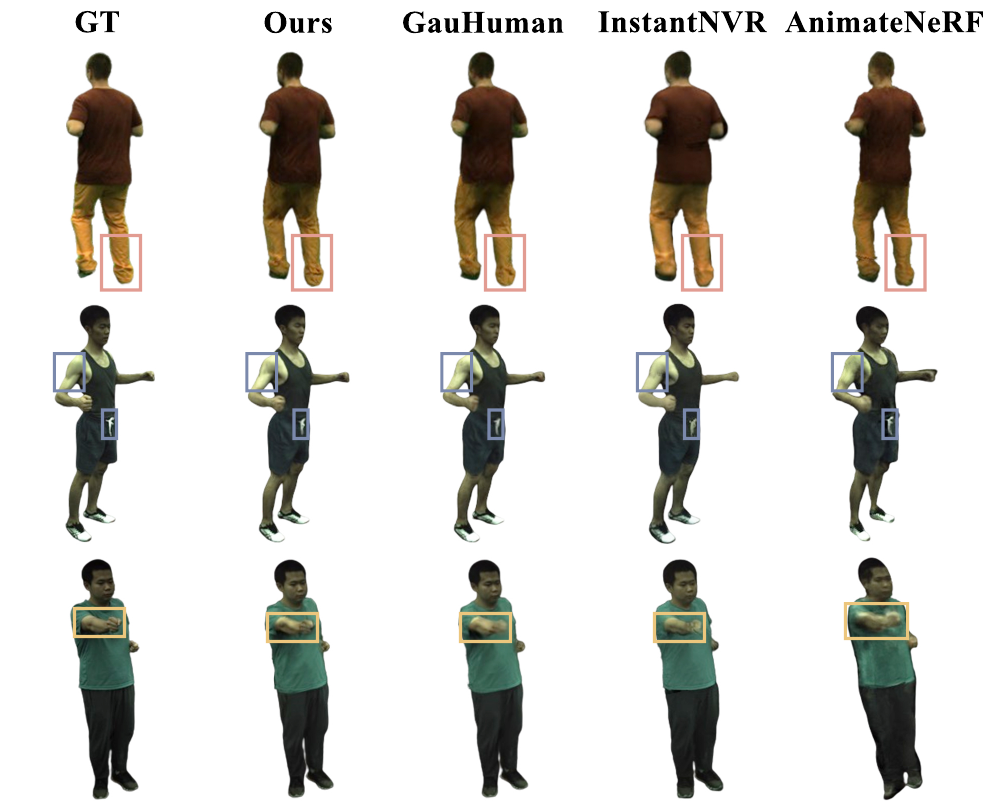}
    \caption{Results from our method and baseline methods on ZJU-MoCap and MonoCap. Our method has superior rendering quality.}
    \label{figure:main_exp}
\end{figure*}

To address the geometric distortion stemming from semantic ambiguities at the junction of body parts, we conducted a comparative analysis of the reconstruction outcomes between the baseline method and our proposed Semantic Kinematic Topology approach across three distinct scenarios illustrated in Figure~\ref{figure:main_exp}. The experimental findings reveal that our model outperforms the baseline methods by producing superior results not only at the intersections of human \textbf{joints and clothing} but also \textbf{at the junctures within the human body}, \textbf{demonstrating clear edge boundaries}. Additionally, our model exhibits improved capability in managing geometric distortions and accurately restoring the shapes of body parts in the presence of occlusions. Among the compared methods, InstantNVR~\cite{InstantNVR} proposes to decompose the human body into multiple structured geometries, but it relies solely on hierarchical shapes and textures for partitioning, without considering rich human semantic priors and lacking an understanding of human motion characteristics. Although Gauhuman~\cite{gauhuman} introduced the human prior of SMPL,  it fails to learn the topological relationships between different body parts, resulting in blurred joint junctions. AnimateNeRF~\cite{AnimaNerf} introduces the human joint skeleton as a regularized learning mechanism for the linkage effects between bones, but lacks precise modeling of the linkage effects between specific body parts and other body parts, leading to human arm deformities in the rendering results. By learning Gaussian human body semantics, our method captures the topological relationships of different body parts, thereby deeply understanding the \textbf{kinematic characteristics} of these parts. This makes our method more advantageous in dealing with \textbf{semantic confusion} caused by complex motion postures and capable of generating more accurate and natural human body reconstruction results.

To address the quality issues associated with local structural reconstruction resulting from significant surface discontinuities on the human body, we conducted a comparative analysis between the baseline method and our proposed Surface Disentanglement module. The experimental results revealed that, in comparison to the baseline, our reconstruction approach yielded superior outcomes across three distinct scenarios as depicted in Figure~\ref{figure:main_exp}. Our method excels in generating refined local structures that \textbf{capture substantial variations in clothing wrinkles, finger details, and other intricate features}. Gauhuman~\cite{gauhuman} employs supervision based on a two-dimensional pixel level, which often leads to the smoothing out of three-dimensional structural disparities, thereby hindering the accurate capture of local clothing wrinkle details. In contrast, InstantNVR~\cite{InstantNVR} utilizes motion parameterization technology to simulate three-dimensional deformation but predominantly focuses on the global structure of the human body, overlooking the intricate and diverse local structural deformations present. Our method stands out \textbf{by leveraging topology and semantic information to disentangle high-frequency features that represent local structures with significant distinctions from the global feature space of the human body}.Through Gaussian point density and distribution optimization, our approach facilitates the precise reconstruction of local structures on the human body surface.

\subsection{Ablation Study}

\indent To verify the effectiveness of our \textbf{SAG} (\textbf{S}emantic \textbf{A}lignment \textbf{G}aussian) module, \textbf{IJG }(\textbf{I}nter \textbf{J}oint \textbf{G}raph) module, and \textbf{SD}(\textbf{S}urface \textbf{D}isentanglement) module, we perform ablation experiments on the ZJU dataset and Monocap dataset. The quantitative results are shown in Table~\ref{tab:ablation_1}.

\begin{figure}[h]
    \centering
    \includegraphics[width=1\linewidth]{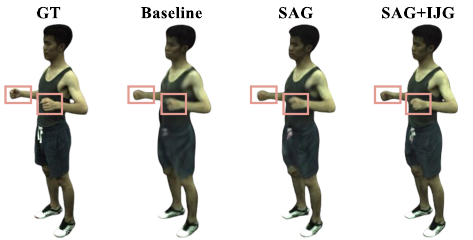}
    \caption{Qualitative results of ablation study on SAG module and IJG module on the 377 sequence of ZJU-Mocap dataset.}
    \label{figure:ab_SAG_IJG}
\end{figure}

\noindent \textbf{Semantic Alignment Gaussian.} We conduct ablation experiments on the 377 sequence of the ZJU-Mocap dataset. The qualitative results are shown in the Figure~\ref{figure:ab_SAG_IJG}. As shown in Figure~\ref{figure:ab_SAG_IJG}, before adding \textbf{SAG}, the Gaussian points fail to be correctly matched to the corresponding geometric spatial positions of the human body parts, resulting in local geometric structure distortion in the rendering results and blurring between parts, as indicated by the boxes in Figure~\ref{figure:ab_SAG_IJG}. After adding the SAG module, our model \textbf{incorporates human body structure information} into the 3D Gaussian optimization process, explicitly indicating specific body part semantics for Gaussian points in the 3D space, so that each body part can independently guide the corresponding Gaussian points for optimization, resulting in clearer rendering results for body parts.

\begin{table*}[t]

\centering
\caption{Quantitative Results of ablating \textbf{S}emantic \textbf{A}lignment \textbf{G}aussian module (\textbf{SAG}), \textbf{I}nter-\textbf{J}oint \textbf{G}raph module (\textbf{IJG}), and \textbf{S}urface \textbf{D}isentanglement module (\textbf{SD}). LPIPS∗ = 1000 × LPIPS}
\tabcolsep=5pt
\resizebox{0.8\textwidth}{!}{
\begin{tabular}{c|c c c |c c c}
\toprule
\multicolumn{1}{c|}{\multirow{2}{*}{Method}} & \multicolumn{3}{c|}{ZJU-Mocap}                 & \multicolumn{3}{c}{MonoCap}             \\
              & PSNR↑     & SSIM ↑   & LPIPS*↓ & PSNR↑  & SSIM ↑ & LPIPS*↓ \\
\midrule
Ours(baseline)    & 28.76     & 0.947    & 31.96   & 31.17  & 0.951  & 16.87 \\
Ours(SAG)     & 30.21     & 0.956    & 30.24   & 32.54  & 0.972  & 14.65 \\
Ours(SAG+IJG)   & 30.87     & 0.960    & 29.52   & 32.77  & 0.980  & 13.14 \\
Our full model(SAG+IJG+SD)          & \textbf{31.35}  & \textbf{0.964}	& \textbf{28.93}	& \textbf{33.61}	& \textbf{0.984}	& \textbf{12.73} \\
\bottomrule
\end{tabular}}
\label{tab:ablation_1}
\end{table*}

\begin{table*}[t]
\centering
\caption{The results of T-Test.}
\tabcolsep=5pt
\resizebox{0.8\textwidth}{!}{
\begin{tabular}{c|c c c c |c c c c}
\toprule
\multicolumn{1}{c|}{\multirow{2}{*}{Method}} & \multicolumn{4}{c|}{PSNR}                 & \multicolumn{4}{c}{LPIPS*}             \\
              & mean      & standard deviation   & t-value & p-value  & mean      & standard deviation   & t-value & p-value \\
\midrule
Ours    & 32.2478     & 0.013    & \multirow{2}{*}{\centering 2.444}   & \multirow{2}{*}{\centering 0.025}  & 17.390  & 0.872 & \multirow{2}{*}{\centering 4.650} & \multirow{2}{*}{\centering 0.0001} \\
GauHuman     & 32.2327     & 0.011    &     &    & 18.924  & 0.913 & & \\
\bottomrule
\end{tabular}}
\label{tab:6}
\end{table*}

To validate the optimal number of categories in the k-nearest neighbors clustering algorithm, we conducted ablation experiments. Table~\ref{tab:para_1} shows the results of the ablation experiment on parameter $k$ mentioned in Equation~\ref{eq:3}. This parameter is the number of different clusters of the human topological graph. Table~\ref{tab:para_1}shows that our model performs the best when $k$=3.

\renewcommand{\arraystretch}{1} 
\begin{table}[h]
\begin{center}
  \caption{Quantitative Results of parameter $k$}
  \label{tab:para_1}
  \resizebox{0.35\textwidth}{!}{
    \begin{tabular}{cccc}
      \toprule
    $k$& PSNR↑  &SSIM↑ &LPIPS↓\\
    \midrule
    2 & 31.54 &0.9681 & 0.01756 \\
    3  & 32.16 & 0.9758 & 0.01738\\
    4  & 31.86 & 0.9673 & 0.01751\\    
    5  & 31.37 & 0.9632 & 0.01764\\    
  \bottomrule
\end{tabular} 
  
  }
   
\end{center}

\end{table}

\begin{figure*}[h]
    \centering
    \includegraphics[width=0.7\linewidth]{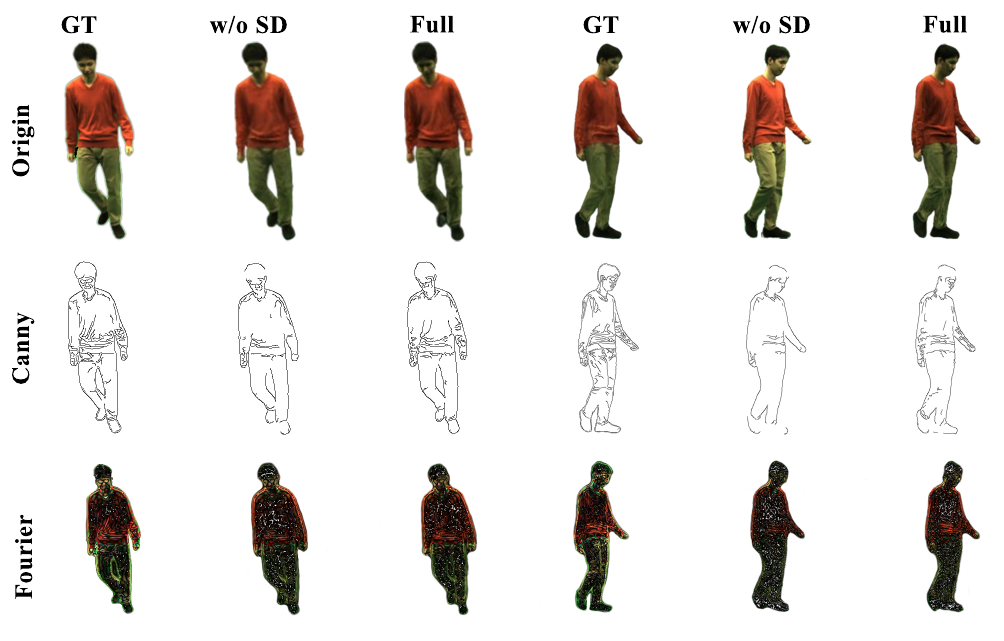}
    \caption{We use the Canny algorithm and Fourier transform to extract the high-frequency regions of the experimental results. The details of the Canny algorithm and Fourier transform can be found in the supplementary section.}
    \label{figure:lan_ab}
\end{figure*}

\textbf{Inter-Joint Graph.} Comparing the model with both SAG and IJG against the model with only SAG, as depicted in Figure~\ref{figure:ab_SAG_IJG}, our IJG model employs a \textbf{semantic topology graph} to leverage the implicit topological relationships of human body parts on surface Gaussian points, resulting in a more accurate reconstruction of the human body. Notably, the representation of the geometric structure at the knees, as indicated by the blue box in Figure~\ref{figure:ab_SAG_IJG}, aligns more closely with the ground truth when employing the IJG model.

\textbf{Surface Disentanglement:} 
Ablation experiments were conducted on the \textbf{S}urface \textbf{D}isentanglement (\textbf{SD}) module to showcase its impact on high-frequency features. The results are presented in Table~\ref{tab:ablation_1}. A comparison with the model without the SD module demonstrates that the complete model can effectively capture high-frequency information on the human body surface. By performing dense operations on Gaussian points in these regions to guide density and attribute optimization, the model enhances the local rendering granularity of the human body surface, presenting high-frequency details more accurately.

To further demonstrate the optimization of the SD module in reconstructing high-frequency features of the human body, we apply the Canny algorithm and Fourier transform to extract the high-frequency regions of the experimental results, which are shown in Figure~\ref{figure:lan_ab}. The model lacking the \textbf{S}urface \textbf{D}isentanglement (\textbf{SD}) module exhibits inadequate responsiveness to high-frequency parts, resulting in incomplete reconstruction of the geometric structure for parts with significant high-frequency information, such as the face and clothing wrinkles. On the other hand, our complete model demonstrates a higher level of restoration for high-frequency features. The SD module performs effectively in restoring high-frequency surface positions, generating more realistic and stable results. This validates the efficacy of the SD module in controlling the density of Gaussian points on the high-frequency surface and optimizing their attributes.

\subsection{T-Test}
We perform ten experiments on the 377 sequences of the ZJU-Mocap dataset, with each experiment using a different camera for training and the remainder for validation.  Identical experiments are performed on both GauHuman and our method because GauHuman ranks second in both PSNR and LPIPS indicators (Table~\ref{tab:comparison}). Through a T-Test, we validate that our method outperforms GauHuman on both PSNR and LPIPS metrics (Table~\ref{tab:6}). We set the significance level at 0.05, where a p-value below this threshold indicates a difference in results between the two methods. The critical value of the T-Test, with 18 degrees of freedom, is determined to be 2.101.

We find that our method performs better in terms of PSNR, with a t-value of 2.444 (greater than the critical value of 2.101), and a p-value\textless 0.05, indicating the correctness of the result. Similarly, on LPIPS, the t-value also exceeds 2.101, and the p-value\textless 0.05, confirming our method's excellent performance.

\section{Conclusion}
We introduce a Human Gaussian Control with Hierarchical Semantic Graphs (HUGS) method for human reconstruction. This method utilizes the Inter-Semantic Kinematic Topology Aware module to learn the relationships in motion at the connections of different body parts and the Intro-Surface Disentanglement module to learn the relationships in appearance within body parts. It addresses blurred details in the reconstruction of the human body, at the junctions and surface features. Our approach explicitly integrates relationships in  motion to reconstruct more intricate human poses and deformations. Through extensive experimentation, we have validated the efficacy of our proposed method, particularly in capturing surface details and articulating body parts. We believe that this innovative approach can provide insights into resolving longstanding reconstruction challenges. Specifically, by integrating semantic and topological information across temporal sequences, it aims to capture the evolving semantic and topological trends in human motion, thereby enabling the utilization of 3DGS in 4D human dynamic modeling. Furthermore, we aim to explore the interplay of topological and semantic relationships between humans and objects to enhance the precision of human reconstruction in complex multi-object interaction scenarios.
\newpage
{   
\small
\bibliographystyle{ieeenat_fullname}
\bibliography{main}
}

\end{CJK}
\end{document}